\documentclass[fleqn,10pt]{wlscirep}
\usepackage[utf8]{inputenc}
\usepackage[T1]{fontenc}
\usepackage{eqnarray,amsmath}
\usepackage{nomencl}
\makenomenclature
\usepackage{multirow}
\usepackage{hhline}

\title{Video Action Recognition Collaborative Learning with Dynamics via PSO-ConvNet Transformer}
\author[1,*]{Nguyen~Huu~Phong}
\author[1]{Bernardete~Ribeiro}
\affil[1]{CISUC, Department of Informatics Engineering, University of Coimbra, Coimbra, Portugal}

\affil[*]{phong@dei.uc.pt}

\begin{abstract}
Recognizing human actions in video sequences, known as Human Action Recognition (HAR), is a challenging task in pattern recognition. While Convolutional Neural Networks (ConvNets) have shown remarkable success in image recognition, they are not always directly applicable to HAR, as temporal features are critical for accurate classification. In this paper, we propose a novel dynamic PSO-ConvNet model for learning actions in videos, building on our recent work in image recognition. Our approach leverages a framework where the weight vector of each neural network represents the position of a particle in phase space, and particles share their current weight vectors and gradient estimates of the Loss function. To extend our approach to video, we integrate ConvNets with state-of-the-art temporal methods such as Transformer and Recurrent Neural Networks. Our experimental results on the UCF-101 dataset demonstrate substantial improvements of up to 9\% in accuracy, which confirms the effectiveness of our proposed method. In addition, we conducted experiments on larger and more variety of datasets including Kinetics-400 and HMDB-51 and obtained preference for Collaborative Learning in comparison with Non-Collaborative Learning (Individual Learning). Overall, our dynamic PSO-ConvNet model provides a promising direction for improving HAR by better capturing the spatio-temporal dynamics of human actions in videos. The code is available at https://github.com/leonlha/Video-Action-Recognition-Collaborative-Learning-with-Dynamics-via-PSO-ConvNet-Transformer.
\end{abstract}
\begin{document}

\flushbottom
\maketitle
%
%
\thispagestyle{empty}
\setlength{\nomlabelwidth}{2.5cm}
\nomenclature{\(X\)}{Input frame in video sequences for ConvNet.}
\nomenclature{\(O_{i}\)}{Output for layer $i_{th}$.}
\nomenclature{\(f_{i}\)}{Weight operation for convolution, pooling or fully connected layers at layer $i_{th}$.}
\nomenclature{\(g_{i}\)}{Activation function at layer $i_{th}$.}
\nomenclature{\(x_{t}\)}{Input sequence of RNN at time step $t$.}
\nomenclature{\(W_{h},W_{x},b,\sigma\)}{Weight matrices, bias, sigmoid function.}
\nomenclature{\(h_{t}\)}{Hidden cell state at time step $t$.}
\nomenclature{\(y_{t},\hat{y}_t\)}{Output of a cell at time step $t$.}
\nomenclature{\(f_{t},i_{t},o_{t},c_{t},c^{'}_{t}\)}{Forget gate, input gate, output gate and cell states at time step $t$.}
\nomenclature{\(x^{n}(t),v^{n}(t)\)}{Position and velocity vector of particle $n$ at time $t$.}
\nomenclature{\(\phi^{(n)}(t),\psi^{(n)}(t)\)}{Intermediate position and intermediate velocity of particle $n$ at time $t$.}
\nomenclature{\(P^{n}(t)\)}{Best position visited up until time $t$ by particle $n$.}
\nomenclature{\(P_{g}^{n}(t)\)}{Best position across all previous positions of the particle $n$ jointly with its nearest-neighbors up until time $t$.}
\nomenclature{\(L\)}{Loss function.}
\nomenclature{\(c,c_{1},c_{2}\)}{Accelerator coefficients.}
\nomenclature{\(r(t)\)}{Random uniform within the interval [0,1].}
\nomenclature{\(M,\beta\)}{Constants.}
\printnomenclature
\section*{Introduction}
Human action recognition plays a vital role for distinguishing a particular behavior of interest in the video. It has critical applications including visual surveillance for detection of suspicious human activities to prevent the fatal accidents~\cite{sultani2018real,li2020abnormal}, automation-based driving to sense and predict human behavior for safe navigation~\cite{razali2021pedestrian,yang2020driver}. In addition, there are large amount of non-trivial applications such as human-machine interaction~\cite{presti20163d,poppe2010survey}, video retrieval~\cite{zhu2017tornado}, crowd scene analysis~\cite{curtis2013right} and identity recognition~\cite{paul2014survey}.
 

In the early days, the majority of research in Human Activity Recognition was conducted using hand-crafted methods~\cite{wang2011action,wang2013action,gorelick2007actions}. However, as deep learning technology evolved and gained increasing recognition in the research community, a multitude of new techniques have been proposed, achieving remarkable results.
 
Action recognition preserves a similar property of image recognition since both of the fields handle visual contents. In addition, action recognition classifies not only still images but also dynamics temporal information from the sequence of images. Built on these intrinsic characteristics, action recognition's methods can be grouped into two main approaches namely recurrent neural networks (RNN) based approach and 3-D ConvNet based approach. Besides of the main ones, there are other methods that utilize the content from both spatial and temporal and coined the name two-stream 2-D ConvNet based approach~\cite{simonyan2014two}.


Initially, action recognition was viewed as a natural extension of image recognition, and spatial features from still frames could be extracted using ConvNet, which is one of the most efficient techniques in the image recognition field. However, traditional ConvNets are only capable of processing a single 2-D image at a time. To expand to multiple 2-D images, the neural network architecture needs to be re-designed, including adding an extra dimension to operations such as convolution and pooling to accommodate 3-D images. Examples of such techniques include C3D~\cite{tran2015learning}, I3D~\cite{carreira2017quo}, R3D~\cite{hara2018can}, S3D~\cite{xie2018rethinking}, T3D~\cite{diba2017temporal}, LTC~\cite{varol2017long}, among others


Similarly, since a video primarily consists of a temporal sequence, techniques for sequential data, such as Recurrent Neural Networks and specifically Long Short Term Memory, can be utilized to analyze the temporal information. Despite the larger size of images, feature extraction is often employed. Long-term Recurrent Convolutional Networks (LRCN)\cite{donahue2015long} and Beyond-Short-Snippets\cite{yue2015beyond} were among the first attempts to extract feature maps from 2-D ConvNets and integrate them with LSTMs to make video predictions. Other works have adopted bi-directional LSTMs~\cite{ullah2017action,he2021db}, which are composed of two separate LSTMs, to explore both forward and backward temporal information.

To further improve performance, other researchers argue that videos usually contain repetitive frames or even hard-to-classify ones which makes the computation expensive. By selecting relevant frames, it can help to improve action recognition performance both in terms of efficiency and accuracy~\cite{gowda2021smart}. A similar concept based on attention mechanisms is the main focus in recent researches to boost overall performance of the ConvNet-LSTM frameworks~\cite{ge2019attention,wu2019adaframe}. 


While RNNs are superior in the field, they process data sequentially, meaning that information flows from one state to the next, hindering the ability to speed up training in parallel and causing the architectures to become larger in size. These issues limit the application of RNNs to longer sequences. In light of these challenges, a new approach, the Transformer, emerged~\cite{vaswani2017attention,touvron2021going,dosovitskiy2020image,arnab2021vivit,liu2022video}.


There has been a rapid advancement in action recognition in recent years, from 3-D ConvNets to 2-D ConvNets-LSTM, two-stream ConvNets, and more recently, Transformers. While these advancements have brought many benefits, they have also created a critical issue as previous techniques are unable to keep up with the rapidly changing pace. Although techniques such as evolutionary computation offer a crucial mechanism for architecture search in image recognition, and swarm intelligence provides a straightforward method to improve performance, they remain largely unexplored in the realm of action recognition.


In our recent research~\cite{phong2022pso}, we developed a dynamic Particle Swarm Optimization (PSO) framework for image classification. In this framework,  each particle navigates the landscape, exchanging information with neighboring particles about its current estimate of the geometry (such as  the gradient of the Loss function) and its position. The overall goal of this framework is to create a distributed, collaborative algorithm that improves the optimization performance by guiding some of the particles up to the best minimum of the loss function. We  extend this framework to action recognition by incorporating state-of-the-art methods for temporal data (Transformer and RNN) with the ConvNet module in an end-to-end training setup.

In detail, we have made the following improvements compared to our previous publication.

We have supplemented a more comprehensive review of the literature on Human Action Recognition. We have implemented the following enhancements and additions to our work:\\
1) We have introduced an improved and novel network architecture that extends a PSO-ConvNet to a PSO-ConvNet Transformer (or PSO-ConvNet RNN) in an end-to-end fashion.\\
2) We have expanded the scope of Collaborative Learning as a broader concept beyond its original application in image classification to include action recognition.\\
3) We have conducted additional experiments on challenging datasets to validate the effectiveness of the modified model.\\
These improvements and additions contribute significantly to the overall strength and novelty of our research.

The rest of the article is organized as follows: In Section~\ref{sec:background}, we discuss relevant approaches in applying Deep Learning and Swarm Intelligence to HAR. In addition, the proposed methods including Collaborative Learning with Dynamic Neural Networks and ConvNet Transformer architecture as well as ConvNet RNN model are introduced in Section~\ref{sec:collab},~\ref{sec:cotrar} and ~\ref{sec:convrnn}, respectively. The results of experiments, the extension of the experiments and discussions are presented in Section~\ref{sec:results}, ~\ref{sec:extension} and~\ref{sec:discussion}. Finally, we conclude our work in Section~\ref{sec:conclusion}.
\section{Related Works}
\label{sec:background}

In recent years, deep learning (DL) has greatly succeed in computer vision fields, e.g., object detection, image classification and action recognition~\cite{ijjina2016human,gowda2021smart,arnab2021vivit}. One consequence of this success has been a sharp increase in the number of investments in searching for good neural network architectures. An emerging promising approach is changing from the manual design to automatic Neural Architecture Search (NAS). As an essential part of 
automated machine learning, NAS automatically generates neural networks which have led to state-of-the-art results~\cite{real2017large,nayman2019xnas,noy2020asap}. Among various approaches for NAS already present in the literature, evolutionary search stands out as one of the most remarkable methods. For example, beginning with just one layer of neural network, the model develops into a competitive architecture that outperforms contemporary counterparts~\cite{real2017large}.  As a result, the efficacy of the their proposed classification system for HAR on UCF-50 dataset was demonstrated~\cite{ijjina2016human} by initializing the weights of a convolutional neural network classifier based on solutions generated from genetic algorithms (GA).

In addition to Genetic Algorithms, Particle Swarm Optimization - a population-based stochastic search method influenced by the social behavior of flocking birds and schooling fish - has proven to be an efficient technique for feature selection~\cite{kennedy1995particle,shi1998modified}. A novel approach that combines a modified Particle Swarm Optimization with Back-Propagation was put forth for image recognition, by adjusting the inertia weight, acceleration parameters, and velocity~\cite{tu2021modpso}. This fusion allows for dynamic and adaptive tuning of the parameters between global and local search capability, and promotes diversity within the swarm. In catfish particle swarm optimization, the particle with the worst fitness is introduced into the search space when the fitness of the global best particle has not improved after a number of consecutive iterations~\cite{chuang2011improved}. Moreover, a PSO based multi-objective for discriminative feature selection was introduced to enhance classification problems~\cite{xue2012particle}.

There have been several efforts to apply swarm intelligence to action recognition from video. One such approach employs a combination of binary histogram, Harris corner points, and wavelet coefficients as features extracted from the spatiotemporal volume of the video sequence~\cite{zhang2022sports}. To minimize computational complexity, the feature space is reduced through the use of PSO with a multi-objective fitness function. 

Furthermore, another approach combining Deep Learning and swarm intelligence-based metaheuristics for Human Action Recognition was proposed~\cite{basak2022union}. Here, four different types of features extracted from skeletal data - Distance, Distance Velocity, Angle, and Angle Velocity - are optimized using the nature-inspired Ant Lion Optimizer metaheuristic to eliminate non-informative or misleading features and decrease the size of the feature set. 

The ideas of applying pure techniques of Natural Language Processing to Computer Vision have been seen in recent years~\cite{arnab2021vivit, dosovitskiy2020image, phong2020rethinking}. By using the sequences of image patches with Transformer, the models ~\cite{dosovitskiy2020image} can perform specially well on image classification tasks. Similarly, the approach was extended to HAR with sequence of frames ~\cite{arnab2021vivit}. In “Video Swin Transformer” ~\cite{liu2022video}, the image was divided into regular shaped windows and utilize a Transformer block to each one. The approach was found to outperform the factorized models in efficiency by taking advantage of the inherent spatiotemporal locality of videos where pixels that are closer to each other in spatiotemporal distance are more likely to be relevant. In our study, we adopt a different approach by utilizing extracted features from a ConvNet rather than using original images. This choice allows us to reduce computational expenses without compromising efficiency, as detailed in Section~\ref{sec:cotrar}.

Temporal Correlation Module (TCM)~\cite{liu2022motion} utilizes fast-tempo and slow-tempo information and adaptively enhances the expressive features, and a Temporal Segment Network (TSN) is introduced to further improve the results of the two-stream architecture~\cite{wang2016temporal}. Spatiotemporal vector of locally aggregated descriptor (ActionS-ST-VLAD) approach designs to aggregate relevant deep features during the entire video based on adaptive video feature segmentation and adaptive segment feature sampling (AVFS-ASFS) in which the key-frame features are selected~\cite{tu2019action}. Moreover, the concept of using temporal difference can be found in the works~\cite{wang2021tdn,jiang2019stm,phong2018action}. Temporal Difference Networks (TDN) approach proposes for both finer local and long-range global motion information, i.e., for local motion modeling, temporal difference over consecutive frames is utilized whereas for global motion modeling, temporal difference across segments is integrated to capture long-range structure~\cite{wang2021tdn}. SpatioTemporal and Motion Encoding (STM) approach proposes an STM block, which contains a Channel-wise SpatioTemporal Module (CSTM) to present the spatiotemporal features and a Channel-wise Motion Module (CMM) to efficiently encode motion features in which a 2D channel-wise convolution is applied to two consecutive frames and then subtracts to obtain the approximate motion representation~\cite{jiang2019stm}. 

Other related approaches that can be mentioned include Zero-Shot Learning, Few-Shot Learning, and Knowledge Distillation Learning~\cite{zhang2022tn,gao2020pairwise,tu2022general}. Zero-Shot Learning and Few-Shot Learning provide techniques for understanding domains with limited data availability. Similar to humans, who can identify similar objects within a category after seeing only a few examples, these approaches enable the model to generalize and recognize unseen or scarce classes. In our proposed approach, we introduce the concept of Collaborative Learning, where particles collaboratively train in a distributed manner.\\

Despite these advances, the field remains largely uncharted, especially with respect to recent and emerging techniques.


%
\section{Proposed Methods}
\label{sec:proposed}
\subsection{Collaborative Dynamic Neural Networks}
\label{sec:collab}
\begin{figure} [hbt]
\begin{center}
\includegraphics[keepaspectratio,width=0.75\textwidth]{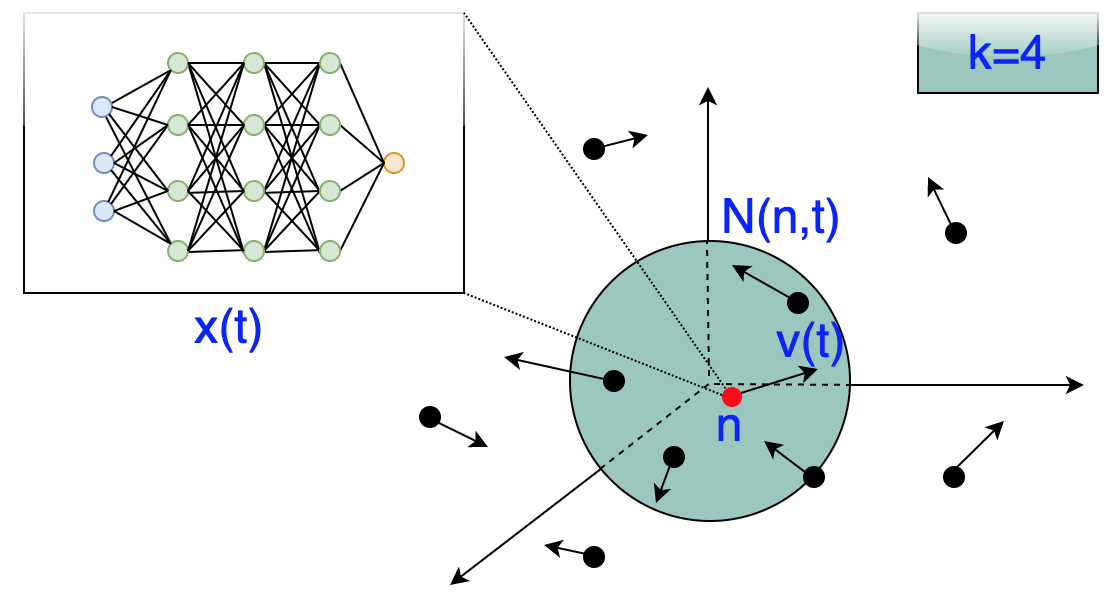}
\caption{A demonstration of the $\mathcal{N}(n,t)$ neighborhood, consisting of the positions of four closest particles and particle $n$ itself, is shown. The velocities of the particles are depicted by arrows.}
\label{fig:nn_concept}
\end{center}
\end{figure}

Define $\mathcal{N}(n,t)$ as the set of $k$ nearest neighbor particles of particle $n$ at time $t$, where $k\in\mathbb{N}$ is some predefined number. In particular, 
\begin{equation}
\begin{aligned}
\mathcal{N}(n,t) ={} & \{(x^{(n)}(t),v^{(n)}(t)),
      (x^{(i_1)}(t),v^{(i_1)}(t)),(x^{(i_2)},v^{(i_2)})(t),\ldots,
      (x^{(i_k)}(t),v^{(i_k)}(t))\}
\end{aligned}
\end{equation}
where $i_1$, $i_2$,... $i_k$ are the $k$ closest particles to $n$ and $x^{(i_k)}(t)$ and $v^{(i_k)}(t)\in \mathbb{R}^D$ represent the position and velocity of particle $i_k$ at time $t$. Figure~\ref{fig:nn_concept} illustrates this concept for $k=4$ particles.

Given a (continuous) function $L\,:\,\mathbb{R}^D\longrightarrow \mathbb{R}$ and a (compact) subset $S\subset \mathbb{R}^D$, define
\begin{equation}
\mathcal{Y}={\sf argmin}\left\{L(y)\,:\,y\in S\right\}
\end{equation}
as the subset of points that minimize $L$ in $S$, i.e., $L(z)\leq L(w)$ for any $z\in \mathcal{Y}\subset S$ and $w\in S$.


\textbf{Dynamic 1:}  
We investigate a set of neural networks that work together in a decentralized manner to minimize a Loss function $L$. The training process is comprised of two phases: i) individual training of each neural network using (stochastic) gradient descent, and ii) a combined phase of SGD and PSO-based cooperation. The weight vector of each neural network is represented as the position of a particle in a $D$-dimensional phase space, where $D$ is the number of weights. The evolution of the particles (or neural networks) is governed by equation ~\eqref{eq:f1}, with the update rule specified by the following dynamics:
\begin{equation}
\begin{array}{ccl}
\psi^{(n)}(t+1) & = & -\eta \nabla L\left(x^{(n)}(t)\right)\\ 
& & \\
\phi^{(n)}(t+1) & = & x^{(n)}(t)+\psi^{(n)}(t+1)\\
& & \\
v^{(n)}(t+1) \!\!\! & \!\!\! = \!\!\! & \!\!\! \sum\limits_{\ell \in \mathcal{N}(n,t)} w_{n\ell} \psi^{(\ell)}(t+1)
+ c_1 r(t)\left(P^{(n)}(t)-\phi^{(n)}(t+1)\right)
+ c_2 r(t)\left(P_g^{(n)}(t)-\phi^{(n)}(t+1)\right)\\
x^{(n)}(t+1) & = & x^{(n)}(t)+v^{(n)}(t)
\end{array} 
\label{eq:f1}
\end{equation}
where $v^{(n)}(t)\in\mathbb{R}^{D}$ is the velocity vector of particle $n$ at time $t$; $\psi^{(n)}(t)$ is an intermediate velocity computed from the gradient of the Loss function at $x^{(n)}(t)$; $\phi^{(n)}(t)$ is the intermediate position computed from the intermediate velocity $\psi^{(n)}(t)$; $r(t)\overset{i.i.d.}\sim {\sf Uniform}\left(\left[0,1\right]\right)$ is randomly drawn from the interval $\left[0,1\right]$ and we assume that the sequence $r(0)$, $r(1)$, $r(2)$, $\ldots$ is i.i.d.; $P^{(n)}(t)\in\mathbb{R}^D$ represents the \emph{best} position visited up until time $t$ by particle $n$, i.e., the position with the minimum value of the Loss function over all previous positions $x^{(n)}(0),\,x^{(n)}(1),\,\ldots,\,x^{(n)}(t)$; $P_{g}^{(n)}(t)$ represents its nearest-neighbors' counterpart, i.e., the best position across all previous positions of the particle $n$ jointly with its corresponding nearest-neighbors~$\bigcup_{s\leq t} \mathcal{N}\left(n,s\right)$ up until time $t$: 
\begin{equation}\label{eq:PSO}
\begin{array}{ccl}
P^{(n)}(t+1) & \in & {\sf argmin}\left\{L(y)\,:\,y=P^{(n)}(t),x^{(n)}(t)\right\}  \\
& & \\
P_{g}^{(n)}(t+1) & \in & {\sf argmin}\left\{L(y)\,:\,y=P_{g}^{(n)}(t),x^{(k)}(t);\right. \\
& & \left.k\in \mathcal{N}(n,t)\right\} \\
& &\\
\end{array}.
\end{equation}
The weights $w_{n\ell}$ are defined as
\begin{equation}
w_{n\ell}= f\left(\left|\left|x^{(n)}(t)-x^{(\ell)}(t)\right|\right|\right),
\end{equation}
with $\left|\left|\cdot\right|\right|$ being the Euclidean norm and $f\,:\,\mathbb{R}\rightarrow \mathbb{R}$ being a decreasing (or at least non-increasing) function. In \underline{Dynamic 1}, we assume that
\begin{equation}
f(z)= \frac{M}{\left(1+z\right)^{\beta}},
\end{equation}
for some constants $M,\beta>0$. This \emph{strengthens} the collaboration learning between any of two particles by pushing each particle against each other.

%

\textbf{Dynamic 2:} An alternative to equation~\eqref{eq:f1} is to pull back a particle instead of pushing it in the direction of the gradient. In the previous section, the assumption was that all particles were located on the same side of a valley in the loss function. However, if one particle is on the opposite side of the valley relative to the rest of the particles, it will be pulled further away from the minimum using the first dynamic. To address this issue, we introduce a second dynamic (Dynamic 2) that pulls the particle back. The formula for this dynamics is as follows:

\begin{equation}
\begin{array}{ccl}
x_{(i)}(t+1) & = & x_{(i)}(t)
+ \sum_{j=1}^N \frac{M_{ij}}{(1+\left|\left|x_i(t)-x_j(t)\right|\right|^2)^\beta} (x_j(t)
- \nabla L(x_j(t)))
+ c r\left(P_{nbest(i)}(t)-x_{i}(t)\right) \\
\end{array} 
\label{eq:f2}
\end{equation}
where $x_{(i)}(t)\in\mathbb{R}^{D}$ is the position of particle $i$ at time $t$; $M$, $\beta$ and $c$ are constants set up by experiments with $\left|\left|\cdot\right|\right|$ being the Euclidean norm; $r(t)\overset{i.i.d.}\sim {\sf Uniform}\left(\left[0,1\right]\right)$ randomly drawn from the interval $\left[0,1\right]$ and we assume that the sequence $r(0)$, $r(1)$, $r(2)$, $\ldots$ is i.i.d.; $P_{nbest(i)}(t)\in\mathbb{R}^D$ represents nearest-neighbors' best , i.e., the best position across all previous positions of the particle $n$ jointly with its corresponding nearest-neighbors~$\bigcup_{s\leq t} \mathcal{N}\left(n,s\right)$ up until time $t$.
\subsection{ConvNet Transformer Architecture for Action Recognition}
\label{sec:cotrar}
\begin{figure}[htb!]
\begin{center}
\includegraphics[keepaspectratio,width=1\textwidth]{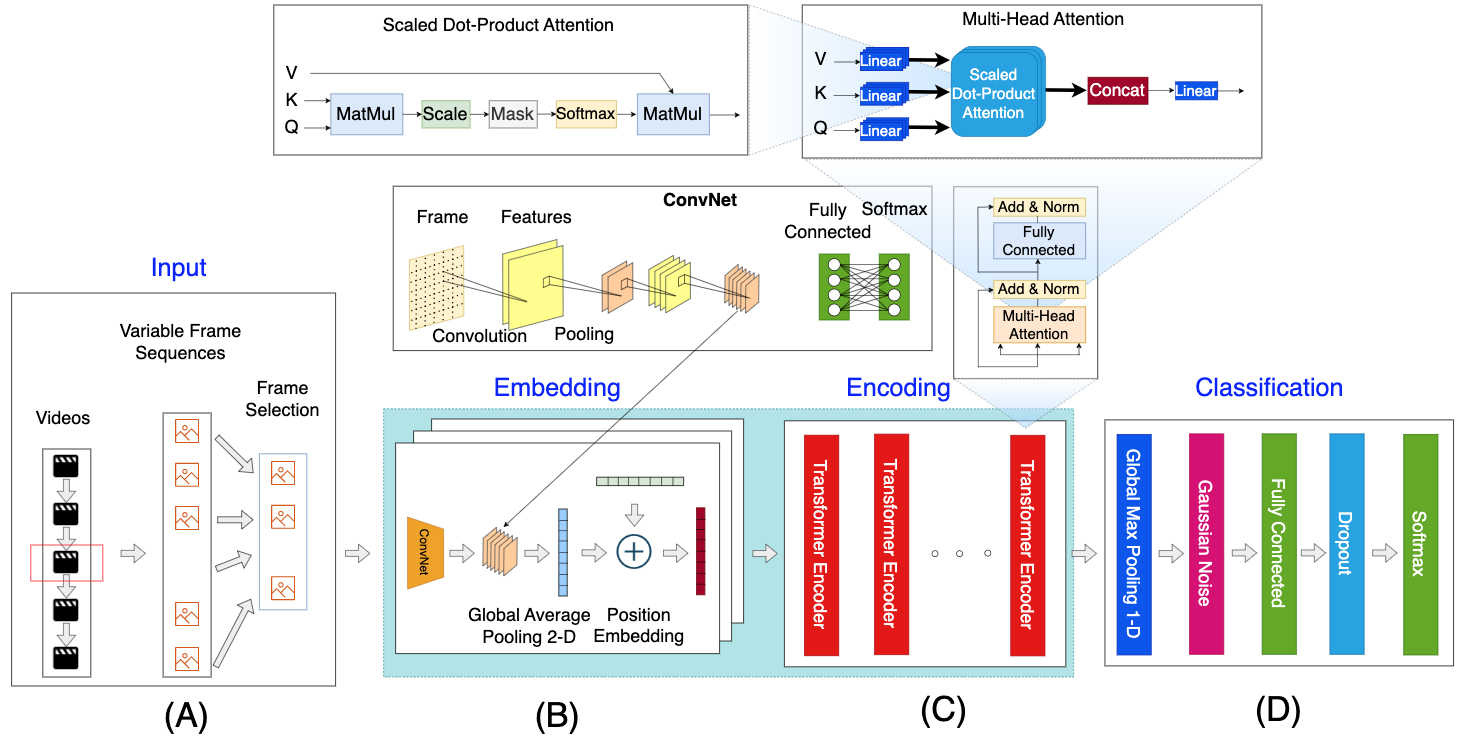}
\caption{Rendering End-to-end ConvNet-Transformer Architecture.}
\label{fig:e2e_cnn_transformer}
\end{center}
\end{figure}

In this section, we discuss a hybrid ConvNet-Transformer architecture that replaces the traditional ConvNet-RNN block for temporal input to classify human action in videos. 
The architecture is composed of several components, including a feature extraction module using ConvNet, a position embedding layer, multiple transformer encoder blocks, and classification and aggregation modules. The overall diagram of the architecture can be seen in Figure~\ref{fig:e2e_cnn_transformer}. The goal of the architecture is to effectively capture the temporal information present in the video sequences, in order to perform accurate human action recognition. The hybrid ConvNet-Transformer design leverages the strengths of both ConvNets and Transformers, offering a powerful solution for this challenging task.
\subsubsection{Features Extraction via ConvNet and Position Embedding} 
In the early days of using Transformer for visual classification, especially for images, the frames were typically divided into smaller patches and used as the primary input~\cite{liu2022video,liu2021swin,zhang2021vidtr}. However, these features were often quite large, leading to high computational requirements for the Transformer. To balance efficiency and accuracy, ConvNet can be utilized to extract crucial features from images, reducing the size of the input without sacrificing performance.

We assume that, for each frame, the extracted features from ConNet have a size of ($w$, $h$, $c$) where $w$ and $h$ are the width and height of a 2D feature and $c$ is the number of filters. 
To further reduce the size of the features, global average pooling is applied, reducing the size from $w \times h \times c$ to $c$.

The position encoding mechanism in Transformer is used to encode the position of each frame in the sequence. The position encoding vector, which has the same size as the feature, is summed with the feature and its values are computed using the following formulas. This differs from the sequential processing of data in the RNN block, allowing for parallel handling of all entities in the sequence.
\begin{equation}
\begin{array}{ccl}
PE_{(pos,2i)} & = & sin(pos\slash{10000^{2i\slash{d_{model}}}})\\
PE_{(pos,2i+1)} & = & cos(pos\slash{10000^{2i\slash{d_{model}}}})
\end{array} 
\label{eq:positionencoding}
\end{equation}
where $pos$, $i$ and $PE$ are the time step index of the input vector, the dimension and the positional encoding matrix; $d_{model}$ refers to the length of the position encoding vector.
\subsubsection{Transformer Encoder}

The Transformer Encoder is a key component of the hybrid ConvNet-Transformer architecture. It consists of a stack of $N$ identical layers, each comprising multi-head self-attention and position-wise fully connected feed-forward network sub-layers. To ensure the retention of important input information, residual connections are employed before each operation, followed by layer normalization.


The core of the module is the multi-head self-attention mechanism, which is composed of several self-attention blocks. This mechanism is similar to RNN, as it encodes sequential data by determining the relevance between each element in the sequence. It leverages the inherent relationships between frames in a video to provide a more accurate representation. Furthermore, the self-attention operates on the entire sequence at once, resulting in significant improvements in runtime, as the computation can be parallelized using modern GPUs.


Our architecture employs only the encoder component of a full transformer, as the goal is to obtain a classification label for the video action rather than a sequence. The full transformer consists of both encoder and decoder modules, however, in our case, the use of only the encoder module suffices to achieve the desired result.

Assuming the input sequence ($X={x_1,x_2,...,x_n}$) is first projected onto these weight matrices $Q = XW_Q$, $K = XW_K$, $V = XW_V$ with $W_Q$,$W_K$ and $W_V$ are three trainable weights, the query ($Q = q_1,q_2,...,q_n$), key ($K = k_1,k_2,...,k_n$) of dimension $d_k$, and value ($V = v_1,v_2,...,v_n$) of dimension $d_v$, the output of self-attention is computed as follows:
\begin{equation}
\begin{array}{ccl}
Attention(Q,K,V)  & = & softmax(\frac{QK^T}{\sqrt{d_k}})V.
\end{array} 
\label{eq:selfattention}
\end{equation}
%
As the name suggested, multi-head attention is composed of several heads and all are concatenated and fed into another linear projection to produce the final outputs as follows:
\begin{eqnarray}
MultiHead(Q,K,V)  & = & Concat(head_1,head_2,...,head_h)W^O. \\
\text{where  } head_i & = &Attention(QW^Q_i,KW^K_i,VW^V_i)
\label{eq:multihead}
\end{eqnarray}
%
%
where parameter matrices $W^Q_i\in\mathbb{R}^{d_{model} \times d_k}$, $W^K_i\in\mathbb{R}^{d_{model} \times d_k}$, $W^V_i\in\mathbb{R}^{d_{model} \times d_v}$ and $W^O\in\mathbb{R}^{hd_v \times d_{model}}$, $i=1,2,...,h$ with $h$ denotes the number of heads.

%
\subsubsection{Frame Selection and Data Pre-processing}

%
Input videos with varying number of frames can pose a challenge for the model which requires a fixed number of inputs. Put simply, to process a video sequence, we incorporated a time distributed layer that requires a predetermined number of frames. To address this issue, we employ several strategies for selecting a smaller subset of frames.

One approach is the "shadow method," where a maximum sequence length is established for each video. While this method is straightforward, it can result in the cutting of longer videos and the loss of information, particularly when the desired length is not reached. In the second method, we utilize a step size to skip some frames, allowing us to achieve the full length of the video while reducing the number of frames used. Additionally, the images are center-cropped to create square images. The efficacy of each method will be evaluated in our experiments.
\subsubsection{Layers for Classification}
Assuming, we have a set of videos $S(S_1,S_2,...,S_m)$ with corresponding labels $y(y_1,y_2,...,y_m)$ where $m$ is the number of samples. We select $l$ frames from the videos and obtain $g$ features from the global average pooling 2-D layer. Each transformer encoder generates a set of representations by consuming the output from the previous block. After $N$ transformer encoder blocks, we can obtain the multi-level representation $H^N(h^N_1,h^N_2,...,h^N_l)$ where each representation is 1-D vector with the length of $g$ (see Figure~\ref{fig:e2e_cnn_transformer} block (A) $\rightarrow$ (D)). 


The classification module incorporates traditional layers, such as fully connected and softmax, and also employs global max pooling to reduce network size. To prevent overfitting, we include Gaussian noise and dropout layers in the design. The ConvNet-Transformer model is trained using stochastic gradient descent and the categorical cross entropy loss is used as the optimization criterion.
\subsection{ConvNet-RNN}
\label{sec:convrnn}


Recent studies have explored the combination of ConvNets and RNNs, particularly LSTMs, to take into account temporal data of frame features for action recognition in videos~\cite{donahue2015long,yue2015beyond,ullah2017action,he2021db,ullah2018action,chen2021lstm}.

To provide a clear understanding of the mathematical operations performed by ConvNets, the following is a summary of the relevant formulations:
\begin{align}
& \begin{cases}
      O_{i}=X & \text{if $i=1$}\\
      Y_{i}=f_{i}(O_{i-1},W_{i}) & \text{if $i>1$}\\
      O_{i}=g_{i}(Y_{i})
  \end{cases} \\
& \begin{cases}
      Y_{i}=W_{i} \circledast O_{i-1} & \text{$i^{th}$ layer is a convolution}\\
      Y_{i}=\boxplus_{n,m} O_{i-1} & \text{$i^{th}$ layer is a pool}\\
      Y_{i}=W_{i}*O_{i-1} & \text{$i^{th}$ layer is a FC}      
  \end{cases}         
\end{align}
where $X$ represents the input image; $O_{i}$ is the output for layer $i^{th}$; $W_{i}$ indicates the weights of the layer; $f_{i}(\cdot)$ denotes weight operation for convolution, pooling or FC layers; $g_{i}(\cdot)$ is an activation function, for example, sigmoid, tanh and rectified linear (ReLU) or more recently Leaky ReLU~\cite{maas2013rectifier}; The symbol ($\circledast$) acts as a convolution operation which uses \textit{shared} weights to reduce expensive matrix computation~\cite{lecun2010convolutional}; Window ($\boxplus_{n,m}$) shows an average or a max pooling operation which computes average or max values over neighbor region of size $n \times m$ in each feature map. Matrix multiplication of weights between layer $i^{th}$ and the layer $(i-1)^{th}$ in FC is represented as ($*$).

The last layer in the ConvNet (FC layer) acts as a classifier and is usually discarded for the purpose of using transfer learning. Thereafter, the outputs of the ConvNet from frames in the video sequences are fed as inputs to the RNN layer.

Considering a standard RNN with a given input sequence ${x_1, x_2,...,x_T}$, the hidden cell state is updated at a time step $t$ as follows:
\begin{equation}
h_t=\sigma(W_h h_{t-1}+W_x x_t+b),
\end{equation}
where $W_h$ and $W_x$ denote weight matrices, b represents the bias, and $\sigma$ is a sigmoid function that outputs values between 0 and 1.

The output of a cell, for ease of notation, is defined as 
\begin{equation}
y_t=h_t,
\end{equation}
but can also be shown using the $softmax$ function, in which $\hat{y}_t$ is the output and $y_t$ is the target:
\begin{equation}
\hat{y_t}=softmax(W_y h_t+b_y).
\end{equation}
A more sophisticated RNN or LSTM that includes the concept of a forget gate can be expressed as shown in the following equations:
\begin{gather}
f_t=\sigma(W_{fh} h_{t-1}+W_{fx} x_t+b_f),\\
i_t=\sigma(W_{ih} h_{t-1}+W_{ix} x_t+b_i),\\
c'_t=tanh(W_{c'h} h_{t-1}+W_{c'x} x_t+b'_c),\\
c_t=f_t \odot c_{t-1}+i_t \odot c'_t,\\
o_t=\sigma(W_{oh} h_{t-1}+W_{ox} x_t+b_o),\\
h_t=o_t \odot tanh(c_t),
\end{gather}
where the $\odot$ operation represents an elementwise vector product, and $f$, $i$, $o$ and $c$ are the forget gate, input gate, output gate and cell state, respectively. Information is retained when the forget gate $f_t$ becomes 1 and eliminated when $f_t$ is set to 0.

For optimization purposes, an alternative to LSTMs, the gated recurrent unit (GRU), can be utilized due to its lower computational demands. The GRU merges the input gate and forget gate into a single update gate, and the mathematical representation is given by the following equations:
\begin{gather}
r_t=\sigma(W_{rh} h_{t-1}+W_{rx} x_t+b_r),
\\
z_t=\sigma(W_{zh} h_{t-1}+W_{zx} x_t+b_z),
\\
h'_t=tanh(W_{h'h} (r_t \odot h_{t-1})+W_{h'x} x_t+b_z),
\\
h_t=(1-z_t) \odot h_{t-1}+z_t \odot h'_t.
\end{gather}

Finally, it's worth noting that while traditional RNNs only consider previous information, bidirectional RNNs incorporate both past and future information in their computations:
\begin{gather}
h_t=\sigma(W_{hx} x_t+W_{hh} h_{t-1} +b_h),
\\
z_t=\sigma(W_{ZX} x_t+W_{HX} h_{t+1} +b_z),
\\
\hat{y_t}=softmax(W_{yh} h_t+W_{yz} z_t+b_y),
\end{gather}
where $h_{t-1}$ and $h_{t+1}$ indicate hidden cell states at the previous time step ($t-1$) and the future time step ($t+1$).
\section{Results}
\label{sec:results}
%
\subsection{Benchmark Datasets}
\label{sec:dataset}

The UCF-101 dataset, introduced in 2012, is one of the largest annotated video datasets available~\cite{soomro2012ucf101}, and an expansion of the UCF-50 dataset. It comprises 13320 realistic video clips collected from YouTube and covers 101 categories of human actions, such as punching, boxing, and walking. The dataset has three distinct official splits (rather than a pre-divided training set and testing set), and the final accuracy in our experiments is calculated as the arithmetic average of the results across all three splits. 

HMDB-51~\cite{kuehne2011hmdb} was released around the same time as UCF-101. The dataset contains roughly 5k videos belonging to 51 distinct action classes. Each class in the dataset holds at least 100 videos. The videos are collected from a multiple sources, for example, movies and online videos.

Kinetics-400~\cite{kay2017kinetics} was recently made available in 2017. The dataset consists of 400 human action classes with at least 400 video clips for each action. The videos were assembled from realistic YouTube in which each clip lasts around 10s. In total, the dataset contains about 240k training videos and 20k validation videos and is one of the largest well-labeled video datasets utilized for action recognition.

Downloading individual videos from the Kinetics-400 dataset poses a significant challenge due to the large number of videos and the fact that the dataset only provides links to YouTube videos. Therefore, we utilize Fiftyone~\cite{fiftyone23}, an open-source tool specifically designed for constructing high-quality datasets, to address this challenge. In our experiment, we collected top-20 most accuracy categories according to the work~\cite{kay2017kinetics} including ``riding mechanical bull'', ``presenting weather forecast'', ``sled dog racing'', etc. Eventually, we obtained 7114 files for training and 773 files for validation with a significant number of files were not collected because the videos were deleted or changed to private, etc. In the same manner, we gathered all categories from HMDB-51 and obtained 3570 files for training and 1530 files for validation. The tool provides one-split for the HMDB-51, but the document does not specify which split.\\

\begin{figure*}[!htb]
\begin{center}
\includegraphics[width=0.75\textwidth]{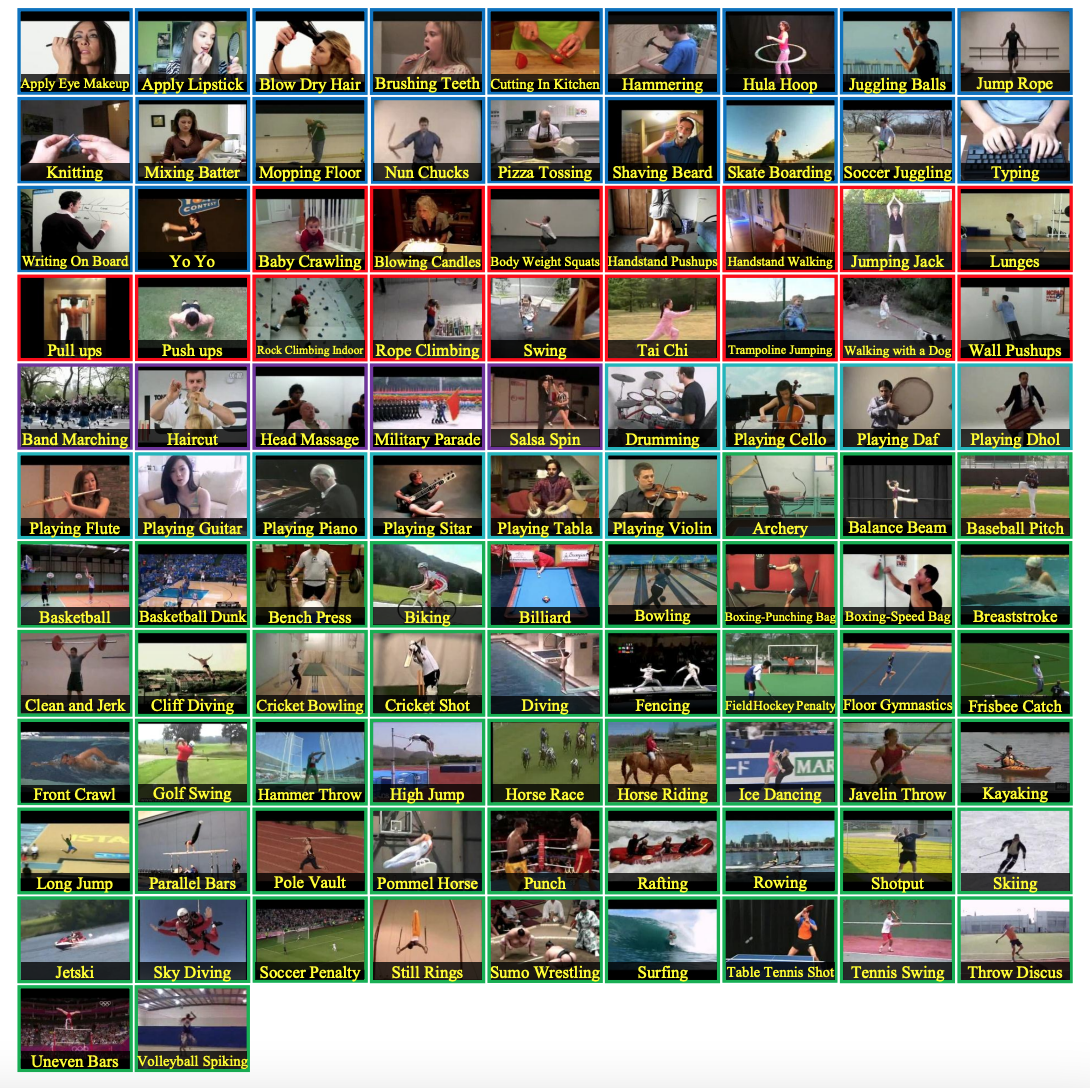}
\caption{A snapshot of samples of all actions from UCF-101 dataset~\cite{soomro2012ucf101}.}
\label{fig:ucf101}
\end{center}
\end{figure*}

Our experiments were conducted using Tensorflow-2.8.2~\cite{abadi2015tensorflow}, Keras-2.6.0, and a powerful 4-GPU system (GeForce\textsuperscript{\textregistered} GTX 1080 Ti). We used Hiplot~\cite{hiplot} for data visualization. Figure~\ref{fig:ucf101} provides snapshot of samples from each of the action categories.
\subsection{Evaluation Metric}
\label{sec:metric}
For evaluating our results, we employ the standard classification accuracy metric, which is defined as follows:
\begin{equation}
Accuracy=\frac{\mbox{Number of correct predictions}}{\mbox{Total numbers of predictions made}}.
\end{equation}
\subsection{Implementation}
\label{sec:implementation}

Training our collaborative models for action recognition involves building a new, dedicated system, as these models require real-time information exchange. To the best of our knowledge, this is the first such system ever built for this purpose. To accommodate the large hardware resources required, each model is trained in a separate environment. After one training epoch, each model updates its current location, previous location, estimate of the gradient of the loss function, and other relevant information, which is then broadcast to neighboring models. To clarify the concept, we provide a diagram of the collaborative system and provide a brief description in this subsection.



\begin{figure*}[htb!]
\begin{center}
\includegraphics[keepaspectratio,width=0.85\textwidth]{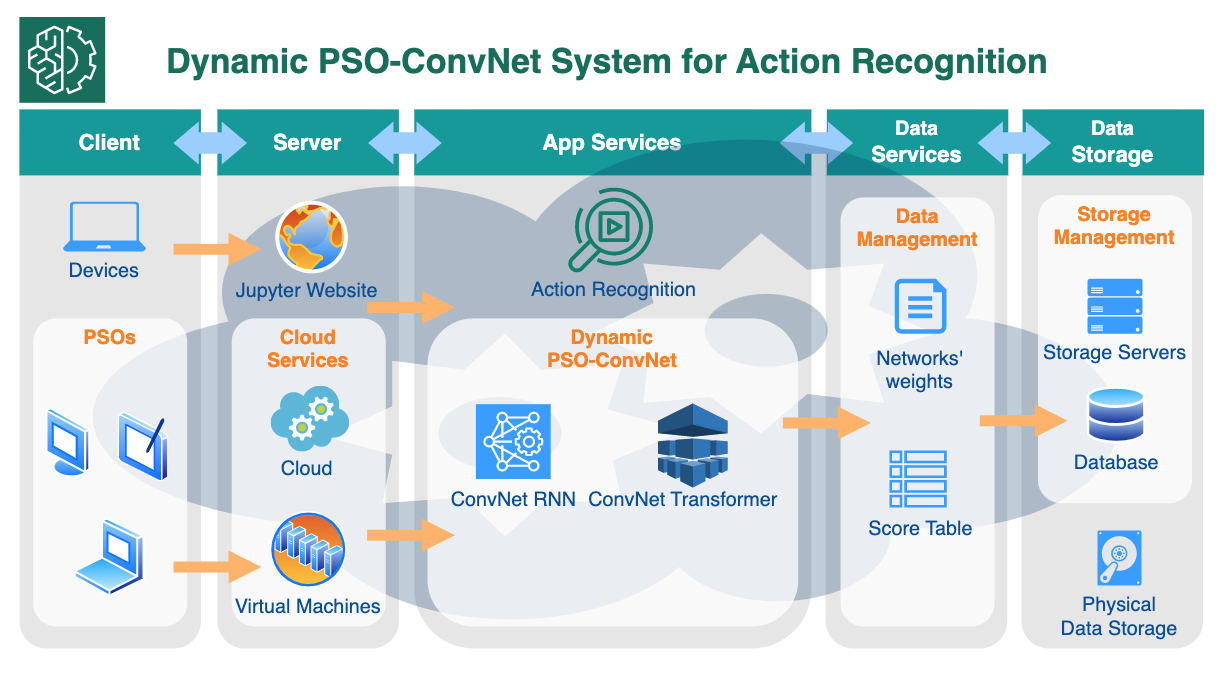}
\caption{Dynamic PSO-ConvNets System Design. The system is divided into two main components, client and server. The client side is accessed through web browser interface while the server side comprises of cloud, app, and data services. The cloud stores virtual machine environments where the models reside. The app service is where the ConvNet-RNN or ConvNet-Transformer runs, and the information generated by each model is managed and saved by the data service. The particles in the system update their positions based on shared information, including current and previous locations, after completing a training cycle.}
\label{fig:dynamic_system}
\end{center}
\end{figure*}

Our system for distributed PSO-ConvNets is designed based on a web client-server architecture, as depicted in Figure~\ref{fig:dynamic_system}. The system consists of two main components: the client side, which is any computer with a web browser interface, and the server side, which comprises three essential services: cloud services, app services, and data services.

The cloud services host the models in virtual machines, while the app services run the ConvNet RNN or ConvNet Transformer models. The information generated by each model is managed by the data services and stored in a data storage. In order to calculate the next positions of particles, each particle must wait for all other particles to finish the training cycle in order to obtain the current information.

The system is designed to be operated through a web-based interface, which facilitates the advanced development process and allows for easy interactions between users and the system.
\subsection{Effectiveness of The Proposed Method}

Table~\ref{tab:table_dynamics_transformer} presents the results of Dynamic 1 and Dynamic 2 on action recognition models. The experiment settings are consistent with our previous research for a fair comparison. As shown in Figure~\ref{fig:e2e_cnn_transformer}, we consider two different ConvNet architectures, namely DenseNet-201 and ResNet-152, and select eight models from the Inception~\cite{szegedy2016rethinking}, EfficientNet~\cite{tan2019efficientnet}, DenseNet~\cite{huang2017densely}, and ResNet~\cite{he2016deep} families. In the baseline action recognition methods (DenseNet-201 RNN, ResNet-152 RNN, DenseNet-201 Transformer, and ResNet-152 Transformer), features are first extracted from ConvNets using transfer learning and then fine-tuned. However, in our proposed method, the models are retrained in an end-to-end fashion. Pretrained weights from the ImageNet dataset~\cite{krizhevsky2012imagenet} are utilized to enhance the training speed. Our results show an improvement in accuracy between $1.58\%$ and $8.72\%$. Notably, the Dynamics 2 for DenseNet-201 Transformer achieves the best result. We also report the time taken to run each method. Fine-tuning takes less time, but the technique can lead to overfitting after a few epochs.

\begin{table}[!htb]
\centering
\caption{Three-fold classification accuracy (\%) on the UCF-101 benchmark dataset. The results of Dynamic 1 and Dynamic 2 using DenseNet-201 Transformer and Resnet-152 Transformer models and compared to baseline models, e.g., Dynamic 1 for DenseNet-201 Transformer versus DenseNet-201 Transformer. The N/A is the abbreviation for the phrase not applicable, for instance, the transfer learning is not applicable in the Dynamic 2 for DenseNet-201 Transformer as the method uses the end-to-end training instead.}
\label{tab:table_dynamics_transformer}
\begin{tabular}{|c|r|c|c|c|c|} 
\hline
\multicolumn{4}{|l|}{}                                                                            & \multicolumn{2}{c|}{Time per fold (h)}             \\ 
\hline
\multicolumn{1}{|l|}{Dynamic Method} & Model                    & Accuracy        & Improve (\%)  & Transfer Learning & Fine-tune/Retrain     \\ 
\hline
\multirow{2}{*}{-}                   & DenseNet-201 Transformer & 0.7741          & N/A             & 26                & \multirow{2}{*}{0.5}  \\ 
\cline{2-5}
                                     & ResNet-152 Transformer   & 0.7679          & N/A             & 24                &                       \\ 
\hline
\multirow{2}{*}{-}                   & DenseNet-201 RNN         & 0.8195          & N/A             & 26                & \multirow{2}{*}{0.1}  \\ 
\cline{2-5}
                                     & ResNet-152 RNN           & 0.8118          & N/A             & 24                &                       \\ 
\hhline{|======|}
\multirow{4}{*}{Dynamic 1}           & DenseNet-201 Transformer & 0.8579          & 8.38          & N/A                 & \multirow{6}{*}{5.5}  \\ 
\cline{2-5}
                                     & ResNet-152 Transformer   & 0.8405          & 7.26          & N/A                 &                       \\ 
\cline{2-5}
                                     & DenseNet-201 RNN         & 0.8462          & 2.67          & N/A                 &                       \\ 
\cline{2-5}
                                     & ResNet-152 RNN           & 0.8276          & 1.58          & N/A                 &                       \\ 
\cline{1-5}
\multirow{2}{*}{Dynamic 2}           & DenseNet-201 Transformer & \textbf{0.8613} & \textbf{8.72} & N/A                 &                       \\ 
\cline{2-5}
                                     & ResNet-152 Transformer   & 0.8399          & 7.20          & N/A                 &                       \\
\hline
\end{tabular}
\end{table}
The experiments described above were conducted using the settings outlined in Table~\ref{tab:table_convnet_settings} and~\ref{tab:table_pso_settings}. The batch size, input image size, and number of frames were adjusted to maximize GPU memory utilization. However, it is worth noting that in Human Activity Recognition (HAR), the batch size is significantly reduced compared to image classification, as each video consists of multiple frames. Regarding the gradient weight M, a higher value indicates a stronger attractive force between particles.\\

\begin{table}[htb]
\centering
\caption{Hyper-parameter settings for the proposed method.}
\label{tab:table_convnet_settings}
\resizebox{\columnwidth}{!}{
\begin{tabular}{l|l|l|l|lllll}
\cline{2-4} \cline{7-9}
                                            & \textbf{Hyper-parameters} & \textbf{Value} & \textbf{Description}       &                       & \multicolumn{1}{l|}{}                        & \multicolumn{1}{l|}{\textbf{Hyper-parameters}} & \multicolumn{1}{l|}{\textbf{Value}} & \multicolumn{1}{l|}{\textbf{Description}}    \\ \cline{1-4} \cline{6-9} 
\multicolumn{1}{|l|}{\textbf{General}}      & BS                        & 8              & Batch size                 & \multicolumn{1}{l|}{} & \multicolumn{1}{l|}{\textbf{Encoding}}       & \multicolumn{1}{l|}{dense\_dim}                 & \multicolumn{1}{l|}{64}             & \multicolumn{1}{l|}{Dense dimension}         \\ \cline{2-4} \cline{7-9} 
\multicolumn{1}{|l|}{\textbf{training}}              & epochs                    & 20             & Number of iterations       & \multicolumn{1}{l|}{} & \multicolumn{1}{l|}{\textbf{layer}}                   & \multicolumn{1}{l|}{num\_heads}                 & \multicolumn{1}{l|}{4}              & \multicolumn{1}{l|}{Number of heads}         \\ \cline{1-4} \cline{6-9} 
\multicolumn{1}{|l|}{\textbf{ConvNet}}      & SIZE                      & (224, 224)     & Input image size           & \multicolumn{1}{l|}{} & \multicolumn{1}{l|}{\textbf{RNN}}            & \multicolumn{1}{l|}{units}                     & \multicolumn{1}{l|}{2048}           & \multicolumn{1}{l|}{Number of memory units}  \\ \cline{2-4} \cline{6-9} 
\multicolumn{1}{|l|}{}                      & CHANNELS                  & 3              & Number of image channels   & \multicolumn{1}{l|}{} & \multicolumn{1}{l|}{\textbf{Classification}} & \multicolumn{1}{l|}{GaussianNoise}             & \multicolumn{1}{l|}{0.1}            & \multicolumn{1}{l|}{Standard deviation of}   \\
\multicolumn{1}{|l|}{}                      &                           &                &                            & \multicolumn{1}{l|}{} & \multicolumn{1}{l|}{\textbf{layer}}                   & \multicolumn{1}{l|}{}                          & \multicolumn{1}{l|}{}               & \multicolumn{1}{l|}{the noise distribution}  \\ \cline{2-4} \cline{7-9} 
\multicolumn{1}{|l|}{}                      & NBFRAME                   & 4              & Number of frames           & \multicolumn{1}{l|}{} & \multicolumn{1}{l|}{}                        & \multicolumn{1}{l|}{Dense}                     & \multicolumn{1}{l|}{1024}           & \multicolumn{1}{l|}{Number of neurons}       \\ \cline{2-4} \cline{7-9} 
\multicolumn{1}{|l|}{}                      & NUM\_FEATURES              & 1920           & Number of features of      & \multicolumn{1}{l|}{} & \multicolumn{1}{l|}{\textbf{}}               & \multicolumn{1}{l|}{Dropout}                   & \multicolumn{1}{l|}{0.4}            & \multicolumn{1}{l|}{Dropout rate}            \\
\multicolumn{1}{|l|}{}                      &                           & or 2048        & DenseNet-201 or ResNet-512 & \multicolumn{1}{l|}{} & \multicolumn{1}{l|}{}                        & \multicolumn{1}{l|}{}                          & \multicolumn{1}{l|}{}               & \multicolumn{1}{l|}{}                        \\ \cline{1-4} \cline{6-9} 
\multicolumn{1}{|l|}{\textbf{Augmentation}} & zoom\_range                & 0.1            & Zoom range                 & \multicolumn{1}{l|}{} & \multicolumn{1}{l|}{\textbf{PSO}}                     & \multicolumn{1}{l|}{num\_neighbors}             & \multicolumn{1}{l|}{4}              & \multicolumn{1}{l|}{Total particles per group}     \\ \cline{2-4} \cline{7-9} 
\multicolumn{1}{|l|}{}                      & rotation\_range            & 8              & Rotation range             & \multicolumn{1}{l|}{} & \multicolumn{1}{l|}{}                        & \multicolumn{1}{l|}{$c_1$}                        & \multicolumn{1}{l|}{0.5}            & \multicolumn{1}{l|}{Coefficient accelerator} \\ \cline{2-4} \cline{7-9} 
\multicolumn{1}{|l|}{}                      & width\_shift\_range         & 0.2            & Width shift range          & \multicolumn{1}{l|}{} & \multicolumn{1}{l|}{}                        & \multicolumn{1}{l|}{$c_2$}                        & \multicolumn{1}{l|}{0.5}            & \multicolumn{1}{l|}{Coefficient accelerator} \\ \cline{2-4} \cline{6-9} 
\multicolumn{1}{|l|}{}                      & height\_shift\_range        & 0.2            & Height shift range         &                       &                                              &                                                &                                     &                                              \\ \cline{2-4}
\multicolumn{1}{|l|}{}                      & Preprocessing      & [-1,1]         & Preprocessing              &                       &                                              &                                                &                                     &                                              \\ \cline{1-4}
\end{tabular}
}
\end{table}
\begin{table}[!htb]
\centering
\caption{Settings of gradient weight $M$.}
\label{tab:table_pso_settings}
\begin{tabular}{ll|cccc|}
\cline{3-6}
                                                &       & \multicolumn{4}{c|}{\textbf{Gradient}}                                                                \\ \cline{3-6} 
                                                &       & \multicolumn{1}{c|}{PSO-1} & \multicolumn{1}{c|}{PSO-2} & \multicolumn{1}{c|}{PSO-3} & PSO-4 \\ \hline
\multicolumn{1}{|l|}{\multirow{4}{*}{\textbf{Gradient}}} & PSO-1 & \multicolumn{1}{c|}{-}     & \multicolumn{1}{c|}{0.2}   & \multicolumn{1}{c|}{0.2}   & 10    \\ \cline{2-6} 
\multicolumn{1}{|l|}{}                          & PSO-2 & \multicolumn{1}{c|}{0.2}   & \multicolumn{1}{c|}{-}     & \multicolumn{1}{c|}{0.2}   & 10    \\ \cline{2-6} 
\multicolumn{1}{|l|}{}                          & PSO-3 & \multicolumn{1}{c|}{0.2}   & \multicolumn{1}{c|}{0.2}   & \multicolumn{1}{c|}{-}     & 10    \\ \cline{2-6} 
\multicolumn{1}{|l|}{}                          & PSO-4 & \multicolumn{1}{c|}{0.2}   & \multicolumn{1}{c|}{0.2}   & \multicolumn{1}{c|}{0.2}   & -     \\ \hline
\end{tabular}
\end{table}
\subsection{Comparison with state-of-the-art methods}
\label{sec:stateoftheart}
\begin{table}[!htb]
\centering
\caption{Comparisons of the proposed method and previous methods on the UCF-101 benchmark dataset. The information of pretrained dataset (if any) are also displayed.}
\label{tab:table_comparepsoactionrecognition}
\resizebox{\columnwidth}{!}{
\begin{tabular}{|l|l|l|c|}
\hline
\textbf{\textbf{Method}}                                               & \textbf{\textbf{Network}} & \textbf{\textbf{Dataset}} & \textbf{\textbf{Accuracy (\%)}} \\ \hline
Bag of words~\cite{soomro2012ucf101} (arXiv'12)                        & -                         & UCF-101                   & 44.5                            \\ \hline
Transfer Learning and Fusions\cite{karpathy2014large} (CVPR'14)        & ConvNet                   & Sports-1M                 & 65.4                            \\ \hline
Shuffle\&Learn~\cite{noroozi2016unsupervised} (ECCV'16)                & AlexNet                   & UCF-101                   & 50.2                            \\ \cline{1-2} \cline{4-4} 
Shuffle\&Learn~\cite{noroozi2016unsupervised} (ECCV'16)                & C3D                       &                           & 55.8                            \\ \hline
TSN RGB image~\cite{wang2016temporal} (ECCV'16)                        & ConvNet                   & UCF-101                   & 84.5                            \\ \cline{1-1} \cline{4-4} 
TSN RGB + Optical Flow + Warped Flow~\cite{wang2016temporal} (ECCV'16) &                           &                           & 92.3                            \\ \hline
DPC~\cite{han2019video} (ICCV'19)                                      & 3D-ResNet34               & Kinetics-400              & 75.7                            \\ \hline
Clip Order~\cite{xu2019self} (CVPR'19)                                 & C3D                       & UCF-101                   & 65.6                            \\ \hline
3D ST-puzzle~\cite{kim2019self} (AAAI'19)                              & C3D                       & Kinetics-400              & 60.6                            \\ \hline
STM~\cite{jiang2019stm} (ICCV'19)                                      & ResNet-50                 & ImageNet+Kinetics         & 96.2                            \\ \hline
P-ODN~\cite{shu2020p} (SR'20)                                          & ConvNet                   & UCF-101                   & 78.6                            \\ \hline
VideoMoCo~\cite{pan2021videomoco} (CVPR'21)                            & R(2+1)D                   & Kinetics-400              & 78.7                            \\ \hline
SVT Slow-Fast~\cite{ranasinghe2022self} (CVPR'22)                                                & TimeSformer               & Kinetics-400 (60K)        & 84.8                            \\ \hline
\textbf{Our model (Dynamics 2 for DenseNet-201 Transformer)}           & DenseNet-201 Transformer  & UCF-101                   & \textbf{86.1}                   \\ \hline
\end{tabular}
}
\end{table}


The comparison between our method (Dynamic 2 for ConvNet Transformer) and the previous approaches is shown in Table ~\ref{tab:table_comparepsoactionrecognition}. The second method (Transfer Learning and Fusions) trains the models on a Sports-1M Youtube dataset and uses the features for UCF-101 recognition. However, the transfer learning procedure is slightly different as their ConvNet architectures were designed specifically for action recognition. While it may have been better to use a pretrained weight for action recognition datasets, such weights are not readily available as the models differ. Also, training the video dataset with millions of samples within a reasonable time is a real challenge for most research centers. Despite these limitations, the use of Transformer and RNN seem to provide a better understanding of temporal characteristics compared to fusion methods. Shuffle\&Learn tries with two distinct models using 2-D images (AlexNet) and 3-D images (C3D) which essentially is series of 2-D images. The accuracy is improved, though, 3-D ConvNets require much more power of computing than the 2-D counterparts. There are also attempts to redesign well-known 2-D ConvNet for 3-D data (C3D is built from scratch of a typical ConvNet~\cite{tran2015learning}), e.g., DPC approach and/or pretrained on larger datasets, e.g., 3D ST-puzzle approach. Besides, VideoMoCo utilizes Contrastive Self-supervised Learning (CSL) based approaches to tackle with unlabeled images. The method extends image-based MoCo framework for video representation by empowering temporal robustness of the encoder as well as modeling temporal decay of the keys. Our Dynamic 2 method outperforms VideoMoCo by roughly $9\%$. SVT is a self-supervised method based on the TimeSformer model that employs various self-attention schemes~\cite{bertasius2021space}. On pre-trained of the entire Kinetics-400 dataset and inference on UCF-101, the SVT achieves 90.8\% and 93.7\% for linear evaluation and fine-tuning settings, respectively. When pre-trained on a subset of Kinetics-400 with $60000$ videos, the accuracy reduces to 84.8\%. Moreover, the TSN methods apply ConvNet and achieves an accuracy of 84.5\% using RGB images (2\% less than our method) and 92.3\% using a combination of three networks (RGB, Optical Flow and Warped Flow). Similarly, the STM approach employs two-stream networks and pre-trained on Kinetics that enhances the performance significantly. Designing a two-stream networks or multi-stream networks would require a larger resource, due to the limitations, we have not pursued this approach at this time. Furthermore, using optical flow~\cite{zhao2022global} and pose estimation~\cite{fang2023alphapose} on original images may improve performance, but these techniques are computationally intensive and time consuming, especially during end-to-end training. The concept of Collaborative Learning, on the other hand, is based on a general formula of the gradient of the loss function and could be used as a plug-and-play module for any approach. Finally, the bag of words method was originally used as a baseline for the dataset and achieved the lowest recognition accuracy ($44.5\%$).

\subsection{Hyperparameter Optimization}

In these experiments, we aimed to find the optimal settings for each model. Table~\ref{tab:table_hybrid} presents the results of the DenseNet-201 Transformer and ResNet-152 Transformer using transfer learning, where we varied the maximum sequence length, number of frames, number of attention heads, and dense size. The number of frames represents the amount of frames extracted from the sequence, calculated by $step=\mbox{(maximum sequence length)/(number of frames)}$. The results indicate that longer sequences of frames lead to better accuracy, but having a large number of frames is not necessarily the best strategy; a balanced approach yields higher accuracy. Furthermore, we discovered that models performed best with 6 attention heads and a dense size of either 32 or 64 neurons.

\begin{figure}[htb!]
\begin{center}
\includegraphics[keepaspectratio,width=0.75\textwidth]{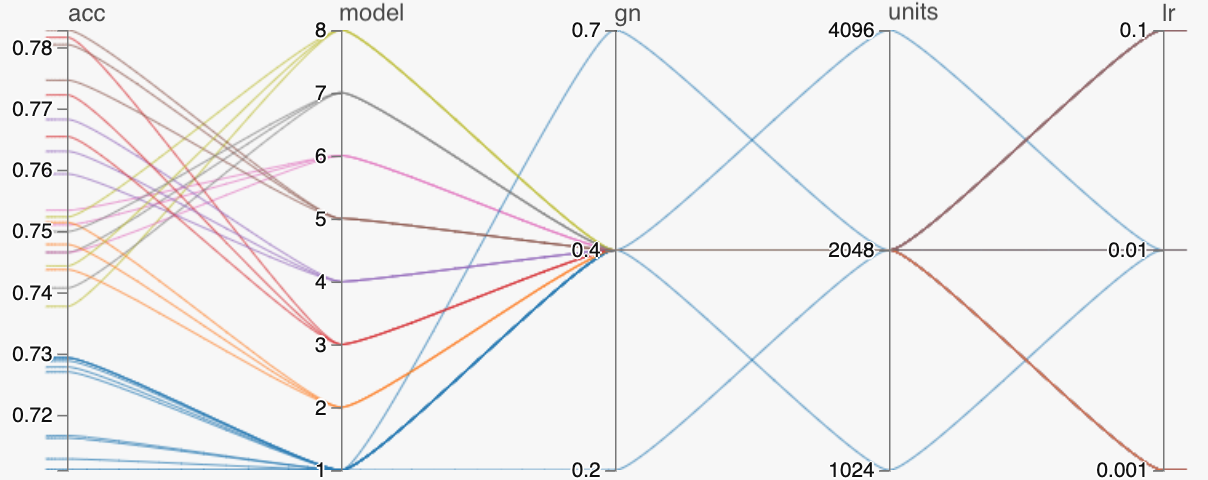}
\caption{Hyperparameter Optimization Results for ConvNet RNN Models with Transfer Learning. The models are numbered as follows: 1.Inception-v3, 2.ResNet-101, 3.ResNet-152, 4.DenseNet-121, 5.DenseNet-201, 6.EfficientNet-B0, 7.EfficientNet-B4, 8.EfficientNet-B7. The abbreviations $acc$, $gn$, and $lr$ stand for accuracy, Gaussian noise, and learning rate, respectively.}
\label{fig:hyperparameter_search}
\end{center}
\end{figure}

\begin{figure*}[htb!]
\begin{center}
\includegraphics[keepaspectratio,width=0.65\textwidth]{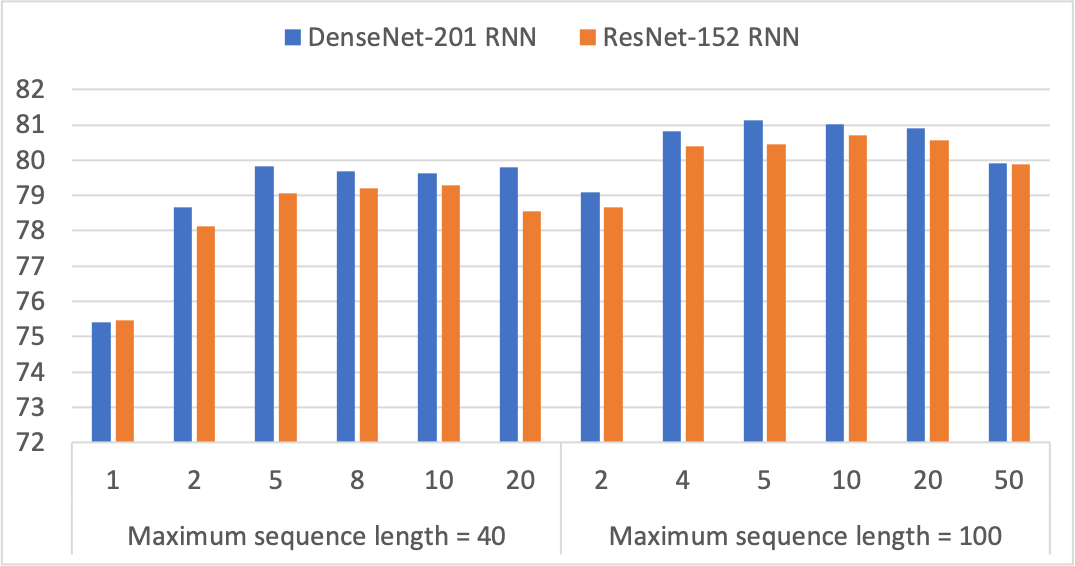}
\caption{Impact of Varying the Number of Frames on the Three-Fold Accuracy of DenseNet-201 RNN and ResNet-152 RNN Using Transfer Learning on the UCF-101 Benchmark Dataset.}
\label{fig:number_of_frames}
\end{center}
\end{figure*}

Figures~\ref{fig:hyperparameter_search} and ~\ref{fig:number_of_frames} show the results for ConvNet RNN models using transfer learning. In the experiments, we first evaluated the performance of eight ConvNets (Inception-v3, ResNet-101, ResNet-152, DenseNet-121, DenseNet-201, EfficientNet-B0, EfficientNet-B4, and EfficientNet-B7). The two best performers, DenseNet-121 and ResNet-152 ConvNet architectures, were selected for further experimentation. The results of varying the number of frames showed a preference for longer maximum sequence lengths.

\begin{table*}
\centering
\caption{Three-Fold Classification Accuracy Results (\%) on the UCF-101 Benchmark Dataset for DenseNet-201 Transformer and ResNet-152 Transformer with Transfer Learning Training.}
\label{tab:table_hybrid}
\resizebox{\columnwidth}{!}{
\begin{tabular}{|c|c|c|c|c|c|c|c|c|} 
\hline
\multicolumn{5}{|c|}{}                                                                                                     & \multicolumn{4}{c|}{Accuracy (\%)}  \\ 
\hline
Model                                      & Maximum sequence length        & Number of frames        & Number of Attention Heads & Dense Size & Set 1 & Set 2 & Set 3 & Avg    \\ 
\hline
\multirow{34}{*}{\rotatebox{90}{DenseNet-201 Transformer}} & \multirow{23}{*}{100} & \multirow{6}{*}{2}  & \multirow{2}{*}{1} & 4          & 69.94 & 69.42 & 69.21 & 69.52  \\ 
\cline{5-9}
                                           &                       &                     &                    & 128        & 69.57 & 69.68 & 68.51 & 69.25  \\ 
\cline{4-9}
                                           &                       &                     & \multirow{2}{*}{4} & 64         & 74.36 & 75.39 & 75.43 & 75.06  \\ 
\cline{5-9}
                                           &                       &                     &                    & 128        & 76    & 75.15 & 74.81 & 75.32  \\ 
\cline{4-9}
                                           &                       &                     & \multirow{2}{*}{8} & 128        & 74.57 & 74.75 & 74.65 & 74.66  \\ 
\cline{5-9}
                                           &                       &                     &                    & 256        & 74.94 & 74.75 & 75.05 & 74.91  \\ 
\cline{3-9}
                                           &                       & \multirow{6}{*}{4}  & \multirow{6}{*}{1} & 4          & 70.92 & 69.47 & 71.67 & 70.69  \\ 
\cline{5-9}
                                           &                       &                     &                    & 8          & 70.02 & 69.63 & 70.32 & 69.99  \\ 
\cline{5-9}
                                           &                       &                     &                    & 16         & 69.87 & 68.53 & 68.29 & 68.90  \\ 
\cline{5-9}
                                           &                       &                     &                    & 32         & 70.37 & 71    & 69.1  & 70.16  \\ 
\cline{5-9}
                                           &                       &                     &                    & 64         & 70.08 & 69.52 & 67.99 & 69.20  \\ 
\cline{5-9}
                                           &                       &                     &                    & 128        & 69.26 & 69.42 & 70.54 & 69.74  \\ 
\cline{3-9}
                                           &                       & \multirow{7}{*}{10} & \multirow{2}{*}{4} & 64         & 76.71 & 76    & 75.3  & 76.00  \\ 
\cline{5-9}
                                           &                       &                     &                    & 128        & 77.43 & 75.33 & 76.46 & 76.41  \\ 
\cline{4-9}
                                           &                       &                     & \multirow{5}{*}{6} & 8          & 77.08 & 76.57 & 76.95 & 76.87  \\ 
\cline{5-9}
                                           &                       &                     &                    & 16         & 77.64 & 76.62 & 76.06 & 76.77  \\ 
\cline{5-9}
                                           &                       &                     &                    & 32         & 77.4  & 76.73 & 77.08 & 77.07  \\ 
\cline{5-9}
                                           &                       &                     &                    & 64         & 76.87 & 77.16 & 78.19 & \textbf{77.41}  \\ 
\cline{5-9}
                                           &                       &                     &                    & 1024       & 74.94 & 73.57 & 72.48 & 73.66  \\ 
\cline{3-9}
                                           &                       & \multirow{4}{*}{20} & \multirow{4}{*}{6} & 8          & 76    & 77.34 & 77.25 & 76.86  \\ 
\cline{5-9}
                                           &                       &                     &                    & 32         & 76.71 & 75.99 & 75.3  & 76.00  \\ 
\cline{5-9}
                                           &                       &                     &                    & 64         & 77.16 & 76    & 76.46 & 76.54  \\ 
\cline{5-9}
                                           &                       &                     &                    & 128        & 77.4  & 75.87 & 77.03 & 76.77  \\ 
\cline{2-9}
                                           & \multirow{11}{*}{40}  & \multirow{8}{*}{2}  & \multirow{4}{*}{4} & 32         & 74.65 & 74.53 & 74.3  & 74.49  \\ 
\cline{5-9}
                                           &                       &                     &                    & 64         & 74.94 & 74.26 & 73.54 & 74.25  \\ 
\cline{5-9}
                                           &                       &                     &                    & 128        & 74.46 & 74.29 & 74.46 & 74.40  \\ 
\cline{5-9}
                                           &                       &                     &                    & 1024       & 71.03 & 70.73 & 70.45 & 70.74  \\ 
\cline{4-9}
                                           &                       &                     & \multirow{4}{*}{6} & 16         & 73.38 & 74.48 & 73.97 & 73.94  \\ 
\cline{5-9}
                                           &                       &                     &                    & 32         & 74.39 & 73.43 & 73.97 & 73.93  \\ 
\cline{5-9}
                                           &                       &                     &                    & 64         & 75.34 & 74.37 & 73.57 & 74.43  \\ 
\cline{5-9}
                                           &                       &                     &                    & 128        & 73.96 & 74.21 & 74.03 & 74.07  \\ 
\cline{3-9}
                                           &                       & \multirow{3}{*}{10} & \multirow{3}{*}{6} & 32         & 76.37 & 74.64 & 74.35 & 75.12  \\ 
\cline{5-9}
                                           &                       &                     &                    & 64         & 75.73 & 75.55 & 74.81 & 75.36  \\ 
\cline{5-9}
                                           &                       &                     &                    & 128        & 76.08 & 76    & 75.08 & 75.72  \\ 
\hline
\multirow{11}{*}{\rotatebox{90}{ResNet-154 Transformer}}   & \multirow{11}{*}{100} & \multirow{6}{*}{2}  & \multirow{2}{*}{1} & 4          & 71.08 & 70.65 & 68.59 & 70.11  \\ 
\cline{5-9}
                                           &                       &                     &                    & 128        & 70.05 & 71.51 & 68.89 & 70.15  \\ 
\cline{4-9}
                                           &                       &                     & \multirow{2}{*}{4} & 128        & 75.02 & 76.03 & 74.73 & 75.26  \\ 
\cline{5-9}
                                           &                       &                     &                    & 256        & 74.76 & 75.66 & 75.19 & 75.20  \\ 
\cline{4-9}
                                           &                       &                     & \multirow{2}{*}{6} & 64         & 75.39 & 75.52 & 74.65 & 75.19  \\ 
\cline{5-9}
                                           &                       &                     &                    & 128        & 75.57 & 75.6  & 75.78 & 75.65  \\ 
\cline{3-9}
                                           &                       & \multirow{5}{*}{10} & \multirow{5}{*}{6} & 4          & 76.55 & 76.78 & 75.65 & 76.33  \\ 
\cline{5-9}
                                           &                       &                     &                    & 8          & 77.08 & 76.35 & 76.14 & 76.52  \\ 
\cline{5-9}
                                           &                       &                     &                    & 16         & 76.55 & 76.25 & 76.33 & 76.38  \\ 
\cline{5-9}
                                           &                       &                     &                    & 32         & 77.24 & 76.49 & 76.65 & \textbf{76.79}  \\ 
\cline{5-9}
                                           &                       &                     &                    & 64         & 75.84 & 77.53 & 76.54 & 76.64  \\
\hline
\end{tabular}
}
\end{table*}
\section{Extension}
\label{sec:extension}
\begin{table}[htb]
\caption{Comparison of Collaborative Learning and Individual Learning on classification accuracy (\%). The experiments are performed using Kinetics-400 and HMDB-51 datasets with DenseNet-201 Transformer and ResNet-152 Transformer models. Each experiment is repeated two times. The N/A is the abbreviation for the phrase not applicable, for instance, the Dynamic methods are not applicable in the Individual Learning. The PSO-1, PSO-2, PSO-3 have the learning rates of $10^{-2}$, $10^{-3}$ and $10^{-4}$, respectively whereas PSO-4's learning rate is in the range [$10^{-1}$, $10^{-5}$].}
\label{tab:table_extension}
\resizebox{\columnwidth}{!}{
\begin{tabular}{lll|clll|clllllll|}
\cline{4-15}
                                                    &                                                                                                          &     & \multicolumn{4}{c|}{Individual Learning}                                                                                                                                                                                                                                                                          & \multicolumn{8}{c|}{Collaborative Learning}                                                                                                                                                                                                                                                                                                                                                                                                                                                                                                                                                                                           \\ \cline{4-15} 
                                                    &                                                                                                          &     & \multicolumn{4}{c|}{N/A}                                                                                                                                                                                                                                                                                          & \multicolumn{4}{c|}{Dynamic 1}                                                                                                                                                                                                                                                                                    & \multicolumn{4}{c|}{Dynamic 2}                                                                                                                                                                                                                                                                                    \\ \hline
\multicolumn{1}{|l|}{Dataset}                       & \multicolumn{1}{l|}{Model}                                                                               &     & \multicolumn{1}{c|}{PSO-1}                                                 & \multicolumn{1}{c|}{PSO-2}                                                 & \multicolumn{1}{c|}{PSO-3}                                                 & \multicolumn{1}{c|}{PSO-4}                                                 & \multicolumn{1}{c|}{PSO-1}                                                 & \multicolumn{1}{c|}{PSO-2}                                                 & \multicolumn{1}{c|}{PSO-3}                                                 & \multicolumn{1}{c|}{PSO-4}                                                 & \multicolumn{1}{c|}{PSO-1}                                                 & \multicolumn{1}{c|}{PSO-2}                                                 & \multicolumn{1}{c|}{PSO-3}                                                 & \multicolumn{1}{c|}{PSO-4}                                                 \\ \hline
\multicolumn{1}{|l|}{\multirow{4}{*}{HMDB-51}}      & \multicolumn{1}{l|}{\multirow{2}{*}{\begin{tabular}[c]{@{}l@{}}ResNet-152 \\ Transformer\end{tabular}}}  & Acc. & \multicolumn{1}{c|}{\begin{tabular}[c]{@{}c@{}}43.97\\ ±0.27\end{tabular}} & \multicolumn{1}{c|}{\begin{tabular}[c]{@{}c@{}}48.09\\ ±1.94\end{tabular}} & \multicolumn{1}{c|}{\begin{tabular}[c]{@{}c@{}}39.13\\ ±0.65\end{tabular}} & \multicolumn{1}{c|}{\begin{tabular}[c]{@{}c@{}}37.82\\ ±8.23\end{tabular}} & \multicolumn{1}{c|}{\begin{tabular}[c]{@{}c@{}}42.73\\ ±1.20\end{tabular}} & \multicolumn{1}{c|}{\begin{tabular}[c]{@{}c@{}}51.43\\ ±0.92\end{tabular}} & \multicolumn{1}{c|}{\begin{tabular}[c]{@{}c@{}}51.43\\ ±1.29\end{tabular}} & \multicolumn{1}{c|}{\begin{tabular}[c]{@{}c@{}}49.60\\ ±0.09\end{tabular}} & \multicolumn{1}{c|}{\begin{tabular}[c]{@{}c@{}}44.07\\ ±1.61\end{tabular}} & \multicolumn{1}{c|}{\begin{tabular}[c]{@{}c@{}}50.84\\ ±1.39\end{tabular}} & \multicolumn{1}{c|}{\begin{tabular}[c]{@{}c@{}}51.24\\ ±1.38\end{tabular}} & \multicolumn{1}{c|}{\begin{tabular}[c]{@{}c@{}}51.27\\ ±1.15\end{tabular}} \\ \cline{3-15} 
\multicolumn{1}{|l|}{}                              & \multicolumn{1}{l|}{}                                                                                    & Max & \multicolumn{1}{c|}{44.17}                                                 & \multicolumn{1}{c|}{49.47}                                                 & \multicolumn{1}{c|}{39.59}                                                 & \multicolumn{1}{c|}{43.65}                                                 & \multicolumn{1}{c|}{43.58}                                                 & \multicolumn{1}{c|}{52.09}                                                 & \multicolumn{1}{c|}{52.35}                                                 & \multicolumn{1}{c|}{49.67}                                                 & \multicolumn{1}{c|}{45.22}                                                 & \multicolumn{1}{c|}{51.83}                                                 & \multicolumn{1}{c|}{52.22}                                                 & \multicolumn{1}{c|}{52.09}                                                 \\ \cline{2-15} 
\multicolumn{1}{|l|}{}                              & \multicolumn{1}{l|}{\multirow{2}{*}{\begin{tabular}[c]{@{}l@{}}DenseNet-201\\ Transformer\end{tabular}}} & Acc. & \multicolumn{1}{l|}{\begin{tabular}[c]{@{}l@{}}45.18\\ ±0.78\end{tabular}} & \multicolumn{1}{l|}{\begin{tabular}[c]{@{}l@{}}52.22\\ ±1.56\end{tabular}} & \multicolumn{1}{l|}{\begin{tabular}[c]{@{}l@{}}41.81\\ ±0.82\end{tabular}} & \begin{tabular}[c]{@{}l@{}}38.21\\ ±9.34\end{tabular}                      & \multicolumn{1}{l|}{\begin{tabular}[c]{@{}l@{}}45.74\\ ±0.46\end{tabular}} & \multicolumn{1}{l|}{\begin{tabular}[c]{@{}l@{}}55.23\\ ±0.65\end{tabular}} & \multicolumn{1}{l|}{\begin{tabular}[c]{@{}l@{}}55.75\\ ±0.09\end{tabular}} & \multicolumn{1}{l|}{\begin{tabular}[c]{@{}l@{}}54.64\\ ±0.83\end{tabular}} & \multicolumn{1}{l|}{\begin{tabular}[c]{@{}l@{}}44.56\\ ±0.83\end{tabular}} & \multicolumn{1}{l|}{\begin{tabular}[c]{@{}l@{}}52.67\\ ±1.29\end{tabular}} & \multicolumn{1}{l|}{\begin{tabular}[c]{@{}l@{}}52.81\\ ±1.48\end{tabular}} & \begin{tabular}[c]{@{}l@{}}51.89\\ ±1.66\end{tabular}                      \\ \cline{3-15} 
\multicolumn{1}{|l|}{}                              & \multicolumn{1}{l|}{}                                                                                    & Max & \multicolumn{1}{l|}{45.74}                                                 & \multicolumn{1}{l|}{\textbf{53.33}}                                                 & \multicolumn{1}{l|}{42.4}                                                  & 44.82                                                                      & \multicolumn{1}{l|}{46.07}                                                 & \multicolumn{1}{l|}{55.69}                                                 & \multicolumn{1}{l|}{\textbf{55.82}}                                                 & \multicolumn{1}{l|}{55.23}                                                 & \multicolumn{1}{l|}{45.15}                                                 & \multicolumn{1}{l|}{53.59}                                                 & \multicolumn{1}{l|}{\textbf{53.86}}                                                 & 53.07                                                                      \\ \hline
\multicolumn{1}{|l|}{\multirow{2}{*}{Kinetics-400}} & \multicolumn{1}{l|}{\multirow{2}{*}{\begin{tabular}[c]{@{}l@{}}DenseNet-201\\ Transformer\end{tabular}}} & Acc & \multicolumn{1}{l|}{\begin{tabular}[c]{@{}l@{}}92.37\\ ±0.45\end{tabular}} & \multicolumn{1}{l|}{\begin{tabular}[c]{@{}l@{}}94.59\\ ±0.27\end{tabular}} & \multicolumn{1}{l|}{\begin{tabular}[c]{@{}l@{}}90.32\\ ±0.12\end{tabular}} & \begin{tabular}[c]{@{}l@{}}86.71\\ ±9.94\end{tabular}                      & \multicolumn{1}{l|}{\begin{tabular}[c]{@{}l@{}}92.5\\ ±0.09\end{tabular}}  & \multicolumn{1}{l|}{94.66}                                                 & \multicolumn{1}{l|}{\begin{tabular}[c]{@{}l@{}}94.72\\ ±0.28\end{tabular}} & \multicolumn{1}{l|}{\begin{tabular}[c]{@{}l@{}}93.42\\ ±1.5\end{tabular}}  & \multicolumn{1}{l|}{\begin{tabular}[c]{@{}l@{}}92.57\\ ±0.55\end{tabular}} & \multicolumn{1}{l|}{\begin{tabular}[c]{@{}l@{}}94.85\\ ±0.82\end{tabular}} & \multicolumn{1}{l|}{\begin{tabular}[c]{@{}l@{}}94.92\\ ±0.73\end{tabular}} & \begin{tabular}[c]{@{}l@{}}94.85\\ ±0.64\end{tabular}                      \\ \cline{3-15} 
\multicolumn{1}{|l|}{}                              & \multicolumn{1}{l|}{}                                                                                    & Max & \multicolumn{1}{l|}{92.7}                                                  & \multicolumn{1}{l|}{\textbf{94.79}}                                                 & \multicolumn{1}{l|}{90.41}                                                 & 93.75                                                                      & \multicolumn{1}{l|}{92.57}                                                 & \multicolumn{1}{l|}{94.66}                                                 & \multicolumn{1}{l|}{\textbf{94.92}}                                                 & \multicolumn{1}{l|}{94.53}                                                 & \multicolumn{1}{l|}{92.96}                                                 & \multicolumn{1}{l|}{\textbf{95.44}}                                                 & \multicolumn{1}{l|}{95.44}                                                 & 95.31                                                                      \\ \hline
\end{tabular}
}
\end{table}
In this Section, we extend our experiments to perform on more challenge datasets, i.e., Kinetics-400 and HMDB-51. In our methods, ConvNets are retrained to improve accuracy performance when compared to Transfer Learning~\cite{phong2022pso,phong2020rethinking}, but these processes can take a long time on the entire Kinetics-400 dataset. As a result, we decided to obtain only a portion of the entire dataset in order to demonstrate our concept. As shown in Table~\ref{tab:table_extension}, our main focus in this study is to compare Non-Collaborative Learning (or Individual Learning) and Collaborative Learning approaches. In each experiment, we conduct two repetitions and record both the mean accuracy and the best accuracy (Max) achieved. All settings are the same as in the experiments with UCF-101 dataset. The learning rate range is obtained by running a scan from a low to a high learning rate. As a consequence, the learning rates of particles PSO-1, PSO-2 and PSO-3 are set at $10^{-2}$, $10^{-3}$ and $10^{-4}$, respectively, whereas the learning rates of the wilder particle PSO-4 has a range of $[10^{-5},10^{-1}]$. The results show a preference for the Collaborative Learning methods as the Dynamic 1 and Dynamic 2 outperform the Individual Learning through both datasets, e.g., an improvement of 0.7\% can be seen on Kinetics-400 using DenseNet-201 Transformer. The results obtained in our experiments clearly demonstrate the superiority of our proposed Collaborative Learning approach for video action recognition.
\section{Discussion}
\label{sec:discussion}

The performance of action recognition methods such as ConvNet Transformer and ConvNet RNN is largely dependent on various factors, including the number of attention heads, the number of dense neurons, the number of units in RNN, and the learning rate, among others. Collaborative learning is an effective approach to improve the training of neural networks, where multiple models are trained simultaneously and both their positions and directions, as determined by the gradients of the loss function, are shared. In our previous research, we applied dynamics to ConvNets for image classification and in this study, we extend the concept to hybrid ConvNet Transformer and ConvNet RNN models for human action recognition in sequences of images. We first aim to identify the optimal settings that lead to the highest accuracy for the baseline models. As seen in Table~\ref{tab:table_dynamics_transformer}, the ConvNet Transformer models did not perform as well as the ConvNet RNN models with transfer learning, which could be due to the limited data available for training, as transformers typically require more data than RNN-based models. However, our proposed method, incorporating dynamics and end-to-end training, not only outperforms the baseline models, but also results in the ConvNet Transformer models outperforming their ConvNet RNN counterparts. This can be attributed to the additional data provided to the transformer models through data augmentation and additional noise.

\section{Conclusion}
\label{sec:conclusion}
%
Recognizing human actions in videos is a fascinating problem in the art of recognition, and while Convolutional Neural Networks provide a powerful method for image classification, their application to HAR can be complex, as temporal features play a critical role.

In this study, we present a novel video action recognition framework that leverages collaborative learning with dynamics. Our approach explores the hybridization of ConvNet RNN and the recent advanced method Transformer, which has been adapted from Natural Language Processing for video sequences. The experiments include the exploration of two dynamics models, Dynamic 1 and Dynamic 2. The results demonstrate a round improvement of $2\%-9\%$ in accuracy over baseline methods, such as an $8.72\%$ increase in accuracy for the DenseNet-201 Transformer using Dynamic 2 and a $7.26\%$ increase in accuracy for the ResNet-152 Transformer using Dynamic 1. Our approach outperforms the previous methods, offering significant improvements in video action recognition. 

In summary, our work makes three key contributions: (1) We incorporate Dynamic 1 and Dynamic 2 into a hybrid model that combines ConvNet with two popular sequence modeling techniques - RNN and Transformer. (2) We extend the distributed collaborative learning framework to address the task of human action recognition. (3) We conducted extensive experiments on the challenging datasets including UCF-101, Dynamics-400 and HMDB-51 over a period of 2-3 months to thoroughly evaluate our approach. To validate its effectiveness, we compared our method against state-of-the-art approaches in the field.
\bibliography{videoactionrecognition}

\section*{Acknowledgements}
This research is sponsored by FEDER funds through the programs COMPETE – “Programa Operacional Factores de Competitividade” and Centro2020 – “Centro Portugal Regional Operational Programme”, and by national funds through FCT – “Fundação para a Ciência e a Tecnologia”, under the project UIDB/00326/2020 and UIDP/00326/2020. The support is gratefully acknowledged.

\section*{Author contributions statement}
N.H.P: Conceptualization, Methodology, Coding, Experiments, Validation, Formal analysis, Investigation, Writing - original draft, Writing - review and editing; B.R.: Writing - review and editing, Supervision. All authors have reviewed the manuscript.

\section*{Competing interests} The authors declare that they have no conflicts of interest.

\section*{Additional information}
\textbf{Availability of Data and Materials} The datasets generated and/or analysed during the current study are available in the UCF101 repository, https://www.crcv.ucf.edu/data/UCF101.php. All data generated or analysed during this study are included in this published article [and its supplementary information files].
\end{document}